\documentclass[10pt,twocolumn,letterpaper]{article}

\usepackage{iccv}
\pdfoutput=1

\usepackage{times}
\usepackage{epsfig}
\usepackage{graphicx}
\usepackage{amsmath}
\usepackage{amssymb}
\usepackage{caption}
\usepackage{multirow}
\usepackage[T1]{fontenc}
\usepackage{rotating}
\usepackage[pagebackref,breaklinks,colorlinks=true]{hyperref}

\iccvfinalcopy 

\def\iccvPaperID{4138} 

\ificcvfinal\fi

\begin{document}

\title{
Disentangle then Parse: \\
Night-time Semantic Segmentation with Illumination Disentanglement
}

\author{%
{Zhixiang Wei{$^{1}$\thanks{Equal contribution.}} ~ Lin Chen{$^{1,2*}$} ~ Tao Tu{$^{1}$} ~ Pengyang Ling{$^{1}$} ~ Huaian Chen{$^{1}$\footnotemark[2]} ~ Yi Jin{$^{1}$\footnotemark[2]}} \\
\normalsize
$^{1}$\	University of Science and Technology of China ~~ $^{2}$\,Shanghai AI Laboratory \\
\normalsize
{\tt\small \{zhixiangwei,chlin,tutao9036,lpyang27,anchen\}@mail.ustc.edu.cn, jinyi08@ustc.edu.cn}
}

\twocolumn[{
\maketitle
}]

\renewcommand{\thefootnote}{\fnsymbol{footnote}}
\footnotetext[1]{~indicates equal contributions.}
\footnotetext[2]{~Corresponding authors.}

\begin{abstract}
Most prior semantic segmentation methods have been developed for day-time scenes, while typically underperforming in night-time scenes due to insufficient and complicated lighting conditions. 
In this work, we tackle this challenge by proposing a novel night-time semantic segmentation paradigm, i.e., disentangle then parse (DTP). DTP explicitly 
disentangles night-time images into light-invariant reflectance and light-specific illumination components and then recognizes semantics based on their adaptive fusion. Concretely, the proposed DTP comprises two key components:
1) Instead of processing lighting-entangled features as in prior works, our Semantic-Oriented Disentanglement (SOD) framework enables the extraction of reflectance component without being impeded by lighting, allowing the network to consistently recognize the semantics under cover of varying and complicated lighting conditions.
2) Based on the observation that the illumination component can serve as a cue for some semantically confused regions, we further introduce an Illumination-Aware Parser (IAParser) to explicitly learn the correlation between semantics and lighting, and aggregate the illumination features to yield more precise predictions.
Extensive experiments on the night-time segmentation task with various settings demonstrate that DTP significantly outperforms state-of-the-art methods. Furthermore, with negligible additional parameters, DTP can be directly used to benefit existing day-time methods for night-time segmentation. Code and dataset are available at \url{https://github.com/w1oves/DTP.git}.
\end{abstract}
\begin{figure}[h]
    \centering
    \includegraphics[width=1.0\linewidth]{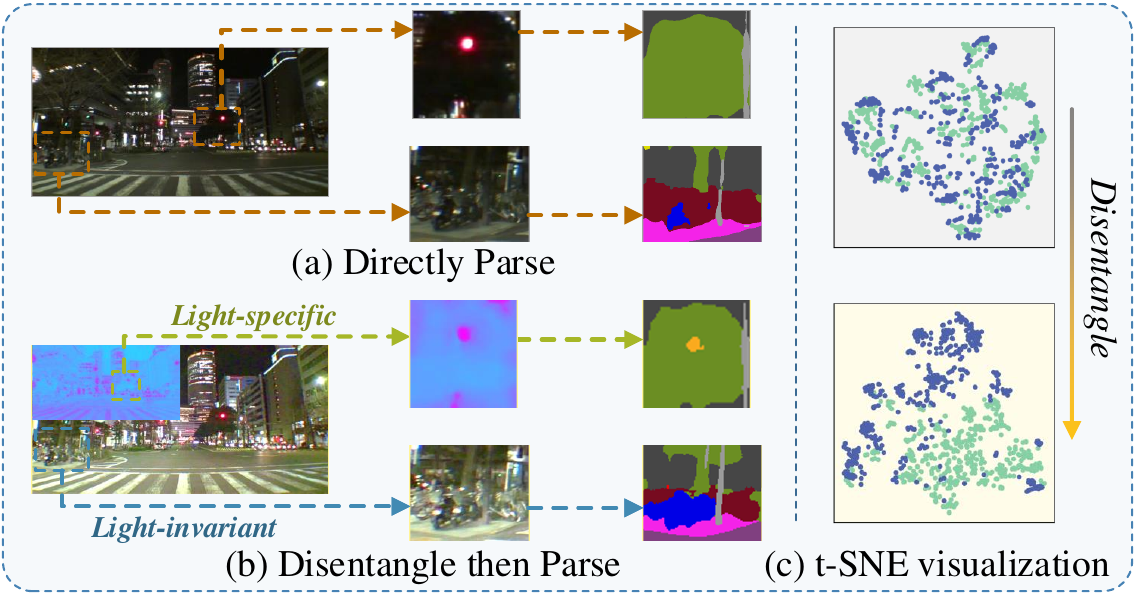}
    \caption{Illustration of the main idea. (a) In the night-time scenes, the entanglement of content and complicated lighting lead to confused semantics. (b) The proposed DTP first disentangles the image and then parses based on the light-invariant and light-specific components. The blue map in the top left corner is a heatmap indicating the lighting intensity. (c) Feature space of road (green) and sidewalk (blue) on the validation set visualized by t-SNE \cite{van2008visualizing} demonstrates that the light-invariant component leads to more discriminative representations.}
    \label{fig:teaser}
\end{figure}
\section{Introduction}
\label{sec:1}
Most existing semantic segmentation methods \cite{lin2017refinenet,chen2018deeplabv3p,Non-Local,PSANet,DANet,OCNet,HANet,strudel2021segmenter,Lawin-Transformer,unet,SSSIPC} are developed for day-time scenes, which have sufficient and uniform lighting. However, in practical application, visual systems are often required to work under insufficient and complicated lighting conditions nearly half the time (\ie, working under the night-time condition), where existing day-time methods may encounter performance drops due to the discrepancy in lighting. 
Therefore, developing a night-time segmentation method, which accounts for the unique characteristics of night-time scenes, is crucial in training a network with stable performance for full-time segmentation.

To achieve similar performance for night-time scenes as for day-time scenes, previous methods \cite{sun2019see,bridging-day-night,dark-adaptation,GPS-GLASS,DDB} have adopted unsupervised domain adaptation techniques to transfer knowledge from labeled day-time domain to unlabeled night-time domain. However, this approach is challenging due to the lack of corresponding labels at night, resulting in limited improvement of segmentation performance. To address this issue, Tan \etal recently propose NightCity \cite{tan2021night}, a large-scale night-time dataset that aims at solving the problem of inadequate training data for night-time segmentation. Several methods have been proposed based on this benchmark, such as EGNet \cite{tan2021night} and NightLab \cite{deng2022nightlab}, which achieve substantial improvement in night-time scenes compared to pure day-time methods. 
However, they typically parse scenes on the lighting-entangled representations, rendering them unsuitable for the challenging lighting conditions in night-time scenes.

Rethinking the challenge of night-time segmentation, we have identified that the main issue lies in insufficient and complex lighting conditions. As shown in Fig.~\ref{fig:teaser} (a), the entanglement of content and complex lighting makes the semantics confused, resulting in blurred object boundaries. 
Therefore, \textbf{an intuitive idea is whether we can disentangle the features of the content and lighting}. 
From this perspective, we propose to disentangle the light-invariant and light-specific components from a night-time image. The light-invariant component allows us to obtain more discriminative features related to the content itself. Fig.~\ref{fig:teaser} (c) illustrates an example of our idea. After separating the light-specific component from the nigh-time image, the feature distributions of the sidewalk and road become more compact and discriminative. Additionally, it is observed that the light-specific component can serve as a cue for some semantically confused regions. As illustrated in Fig.~\ref{fig:teaser} (b), the artificial light usually appears in some special categories (\eg, traffic light and car), which exhibit larger intensity in the light-specific component. 

Based on the above observation, we intend to establish a paradigm that can harness the advantages of light-invariant and light-specific components. However, disentangling these two components is quite challenging due to the lack of well-defined ground truth. Although the Retinex \cite{land1977retinex} theory points out that the light-specific component (\ie, illumination) and light-invariant component (\ie, reflectance) can be organized together via a simple dot product and the light-specific component should satisfy the piece-wise smooth constraint, these weak constraints are inadequate in complex lighting conditions. 
To tackle this challenge, this work employs semantics as an additional constraint for the disentanglement tasks. Specifically, we design a semantic-oriented disentanglement (SOD) framework, which generates training data with ground truth by combining the light-invariant and light-specific components disentangled from different night-time images. Such a training framework allows us to establish semantic consistent constraints for two images sharing the same light-invariant component but owning different light-specific components. Moreover, to make full use of the light-specific component, we propose an illumination-aware parser (IAParser), which explicitly learns the correlation between illumination and semantics, thereby improving the performance of some challenging categories.

Last but not least, we have noticed that NightCity \cite{tan2021night,deng2022nightlab} contains numerous mislabeled pixels, which can potentially hinder the research progress of the night-time segmentation task. In this work, we build a NightCity-fine dataset by elaborately correcting the erroneous labels in NightCity. Supported by this refined dataset, the segmentation model can be evaluated more validly and achieves improved performance. Through extensive experiments, we demonstrate the superiority of the proposed method and the reliability of the refined dataset. In a nutshell, the \textbf{main contributions} of this paper are as follows:
\begin{itemize}
\item We propose a novel night-time semantic segmentation paradigm, \ie disentangle then parse (DTP), to tackle the challenge of insufficient and complicated lighting. With negligible additional parameters, DTP can be readily applied to enhance existing day-time methods for night-time segmentation.
\item We devise a semantic-oriented disentanglement framework (SOD), which disentangles images into light-invariant reflectance and light-specific illumination components with the aid of semantic constraints, allowing the network to extract consistent features under varying lighting. Moreover, we present an illumination-aware parser (IAParser) that harnesses the illumination component to serve as a cue for more precise predictions.
\item We introduce the NightCity-fine dataset by refining the largest night-time segmentation dataset NightCity. Together with DTP, NightCity-fine presents a more robust benchmark for night-time segmentation.
\end{itemize}

\section{Related work}
\label{sec:2}
\begin{figure*}[htbp]
    \centering
    \includegraphics[width=1.0\linewidth]{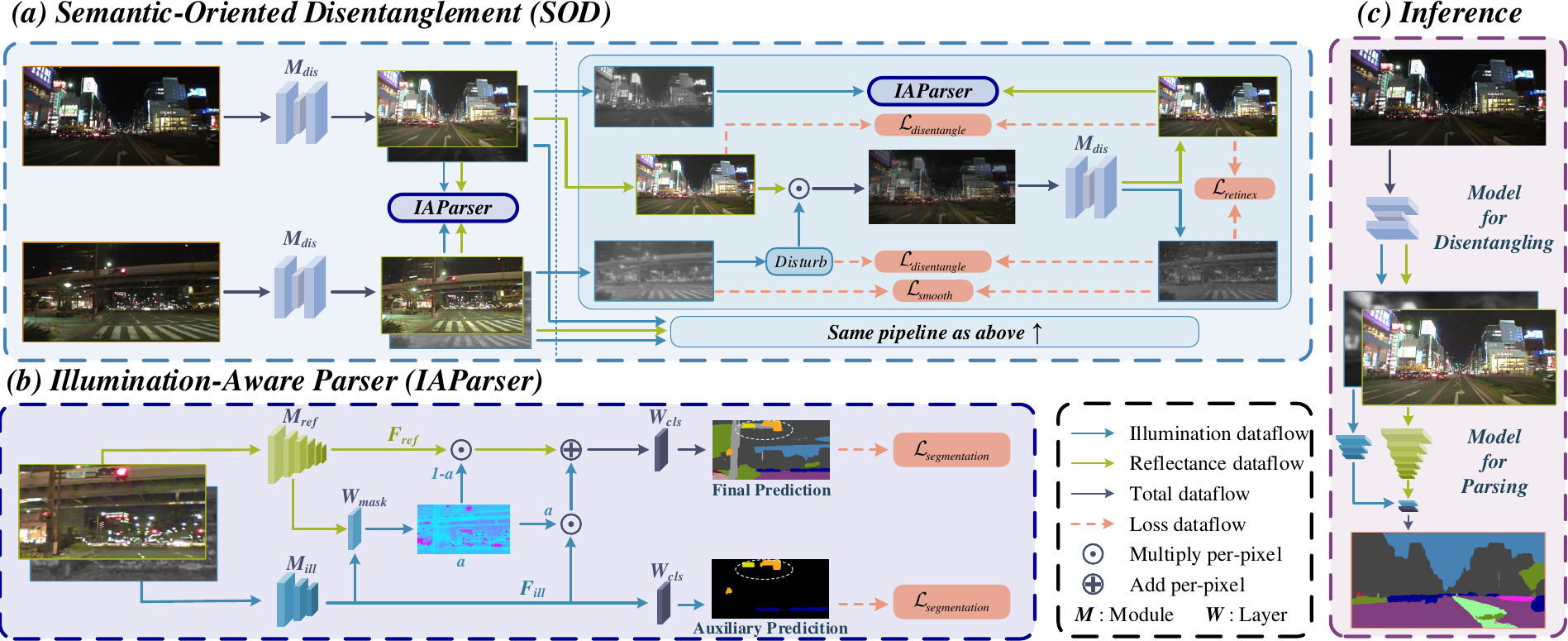}
    \caption{An overview of proposed DTP that consists of SOD and IAParser. In SOD, $M_{dis}$ disentangles the image into two components: light-invariant reflectance and light-specific illumination. In IAParser, an adaptive fusion is adopted to aggregate the features of reflectance and illumination to produce a reliable prediction. DTP can be trained end-to-end and requires only 2M additional parameters to achieve a significant improvement. "disturb" means adding noise to the input illumination, and 'a' is the heatmap where red indicates high intensity.}
    \label{fig:overview}
\end{figure*}
\noindent\textbf{Semantic segmentation}. Semantic segmentation aims to assign each pixel its own category. Later, methods based on Fully Convolutional Networks (FCN)\cite{long2015fully}  combined with encoder-decoder architectures have become the dominant approach for segmentation. Various methods \cite{chen2014deeplab,chen2017deeplabv2,yu2015multi} employ dilated convolutions to enlarge the receptive field, while PSPNet \cite{zhao2017pyramid} leverages pyramid pooling module (PPM) to model multi-scale contexts. Combining these enhancements,  DeepLab series \cite{chen2014deeplab,chen2017deeplabv2,chen2017deeplabv3,chen2018deeplabv3p} propose the atrous spatial pyramid pooling (ASPP) to embed contextual information. More recently, self-attention mechanisms have been widely adopted to capture long-range dependencies for semantic segmentation, such as Non-Local \cite{Non-Local}, DANet \cite{DANet}, and OCRNet \cite{OCRNet}. Furthermore, several transformer-based networks,  such as Vision Transformer \cite{dosovitskiy2020image}, and Swin Transformer \cite{liu2021swin}, have been utilized for stronger backbones. Additionally, ConvNeXt \cite{liu2022convnet} has been presented as a competitive backbone to transformers. However, most of these methods mainly focus on the day-time scene.

\noindent \textbf{Night-time semantic segmentation}. The goal of night-time semantic segmentation is to achieve comparable performance to the day-time counterpart. In the early stage, due to the lack of large-scale labeled datasets in the night-time scenes, some previous methods \cite{bridging-day-night,dark-adaptation,GPS-GLASS,sakaridis2019guided,wu2021dannet} have adopted domain adaptation to transfer the knowledge from the day-time scenes to night-time. Others methods \cite{night-day-trans-dec,night-day-trans-recog} have employed pre-trained image enhancement models during the inference pipeline to translate night-time images into their day-time counterparts. Recently, several methods directly train a model on the labeled night-time data. For instance, Tan \etal propose EGNet \cite{tan2021night}, which uses the V channel of the input image in the HSV color space to guide the network to generate discriminative features in under- and over-exposed regions. Dan \etal propose the NightLab \cite{deng2022nightlab}, in which objects are divided into simple and difficult categories. A Hardness Detection Module is used to detect difficult categories on the fly and send them to a specialized segmentation module. Xie \etal \cite{FDLNet} propose to exploit the image frequency distributions for night-time scene parsing. However, these methods do not explicitly estimate the effect of lighting on semantics but instead implicitly force the network to learn the entangled representations of various content and lighting. 

\noindent\textbf{Disentangling deep representations}. Disentangling the accidental scene events, such as illumination, shadows, viewpoint, and object orientation from the intrinsic scene properties has been a long-desired goal in computer vision \cite{bengio2013representation}. 
This allows the deep-learning models to capture the isolated factors of variation affecting the represented entities, which is crucial in improving their robustness to varying conditions \cite{mathieu2016disentangling,bengio2013representation}.
Previous works \cite{Baslamisli_2018_ECCV} have explored various approaches for disentangling the image representations, such as learning domain-invariant representations across multiple domains in a GAN \cite{goodfellow2020generative} framework\cite{ganin2015unsupervised,bousmalis2017unsupervised,bousmalis2016domain,lee2020drit++,liu2018detach}. In this work, we develop a novel disentangling method for the night-time segmentation task based on the Retinex \cite{land1977retinex} theory and semantic guidance, which significantly improves the performance of the night-time segmentation methods.

\section{Method}
\label{sec:3}
\subsection{Motivation}
\label{sec:3.1}
Night-time scenes are typically characterized by low-intensity lighting and complex artificial light sources, leading to variations in the appearance of objects due to changing lighting conditions. This reinforces the entanglement between light-invariant reflectance and light-specific illumination, making it challenging to extract discriminative features for semantic segmentation. 
Therefore, an intuitive idea is disentangling the illumination from the night-scene image, and thus the consistent features under varying lighting conditions induced by light-invariant reflectance enable robust night-time segmentation. 

Moreover, we have observed that the illumination information can serve as a cue for some semantically confused regions, such as the traffic lights and signs, as illustrated in Fig.~\ref{fig:teaser} (b). Consequently, for a superior segmentation performance, it is necessary to perform an adaptive information fusion of reflectance and illumination based on their semantic characteristics. 

To implement the above idea, we establish a segmentation paradigm as shown in Fig.~\ref{fig:overview}, \ie disentangle then parse, which is formulated as:
\begin{equation}
\label{eq:overview}
    \Tilde{Y}=M_{seg}(R,I)=M_{seg}(M_{dis}(X)),
\end{equation}
where $X$ is an input image, and $\Tilde{Y}$ is the corresponding pixel-wise prediction. The disentanglement module $M_{dis}$ disentangles images into light-invariant reflectance $R$ and light-specific illumination $I$, described in Section \ref{sec:3.2}. Then, the segmentation module $M_{seg}$  incorporates both reflectance and illumination components to produce a reliable prediction, described in Section \ref{sec:3.3}.

\subsection{Semantic-Oriented Disentanglement}
\label{sec:3.2}
According to Retinex theory \cite{land1977retinex}, an image $X$ can be naturally modeled as: 
\begin{equation}
    \label{eq:1}
    X=R\odot I,
\end{equation}
where $R$ represents the light-invariant reflectance that describes the intrinsic property of captured objects, $I$ denotes the light-specific illumination representing various lighting on objects, and $\odot $ is the element-wise multiplication.

Disentangling night-time images solely based on the Retinex theory is a challenging problem in night-time segmentation involving complex lighting conditions. It is difficult to look for a generalized pre-trained disentanglement module, and supervised training for the disentanglement module is not feasible due to the absence of a corresponding dataset. To obtain a disentanglement module that is stable and suitable for night-time segmentation, we utilize the semantic supervision supported in the dataset and innovatively train the disentanglement and segmentation modules together. Specifically, our proposed framework is based on the following two initial points.

\noindent\textbf{Implement disentanglement with entanglement}. Retinex theory assumes that images of the same scene captured in different light conditions share the same reflectance \cite{land1977retinex,wei2018deep}. Based on this assumption, in this work, we first coarsely disentangle night-time images into reflectance and illumination. Then, new synthetic night-time images are entangled with predicted reflectance and various illuminations. Finally, the reconstructed reflectance and illumination are disentangled from synthetic images for calculating pixel-wise reconstruction loss. Through this process, the disentanglement module is guided to learn the consistency of reflectance under different lighting conditions.

\noindent\textbf{Superior disentanglement with semantic constraint}. Reflectance implies the intrinsic properties of categories presented in the scene \cite{nayar1996reflectance}. Therefore, a well-disentangled reflectance should have clearer and more discriminative semantics than an image consisting of incorrect reflectance and redundant illumination. To evaluate the quality of the disentangled reflectance, we propose to train a semantic segmentation module alongside the disentanglement module, which can serve as a discriminator. Specifically, we use the optimization cost of the segmentation predictions as a valid quality evaluation metric for the reflectance.

Following the above guidelines, we design our framework as illustrated in Fig.~\ref{fig:overview} (a). The framework takes two input images as the input, \ie $X_j$ and $X_k$, and feeds them into the disentanglement module $M_{dis}$, which estimates the corresponding reflectance and illumination images, denoted as $R_j$, $I_j$, $R_k$, and $I_k$, respectively. In the initial phase, the estimated reflectance and illumination may be invalid due to inadequate training. In order to obtain meaningful results during this stage, the disentanglement module employs a long-distance jump connection structure (more details are provided in the supplementary material). To prevent the module from learning a fixed transformation, we introduce random noise to disturb illumination:
\begin{equation}
    \label{eq:2}
    I_j^\prime=(1-\beta ) I_j+\beta N_j,I_k^\prime=(1-\beta) I_k+\beta N_k,
\end{equation}
where $\beta\sim Uniform(1-\frac{t}{T},1)$, $t$ is the number of current iteration, and $T$ is the number of maximum iteration. The guidance noise $N_j$ and $N_K$ is designed to possess similar properties and characteristics as common illumination \cite{jiang2021switched,cai2018learning-deep-single,zhao2021retinexdip,ma2021retinexgan,lee2020bright-channel-prior}, which serves as a rough guidance for illumination during the initial training stage. Multiple guidance noise distributions are used together, and more details can be found in the supplementary material.

\begin{table*}[tbp]
\setlength{\abovecaptionskip}{0cm}
\setlength{\belowcaptionskip}{-0.2cm}
\caption{ Comparison results of night-time segmentation in terms of mIoU. The mIoU (NightCity) column  shows the results obtained using a model trained and evaluated on NightCity dataset, while the mIoU (NightCity-fine) column shows the results obtained on NightCity-fine dataset. Results marked with * denote metrics reported in the original article. The best score is indicated in \textbf{bold}.}
\label{tab:nightcity}
\begin{center}
\renewcommand\arraystretch{1.04}
\begin{tabular}{lccccc}
    \hline
    Method&Backbone&Parameters&mIoU (NightCity)&mIoU (NightCity-fine)\\
    \hline    
    NightCity \cite{tan2021night}&Res101&84.6M&\ \ 51.8$^*$&55.9\\
    NightLab \cite{deng2022nightlab}&Swin-Base&242.4M&59.5&62.3\\
    \hline    
    HRNetV2 \cite{HRNetV2}&HRNet-W48&65.8M&58.2&60.3\\
    HRNetV2+\textbf{Ours}&HRNet-W48&68.6M&59.4 (1.2$\uparrow$) &61.8 (1.5$\uparrow$)\\
    \hline
    DLV3P \cite{chen2018deeplabv3p}&Res101&60.1M&54.7&58.8\\    
    DLV3P+\textbf{Ours}&Res101&63.9M&57.6 (2.9$\uparrow$) &60.4 (1.6$\uparrow$) \\
    \hline
    UPer-ConvNeXt \cite{liu2022convnet}&ConvNeXt-Base&120.7M&58.7&60.9\\
    UPer-ConvNeXt + \textbf{Ours}&ConvNeXt-Base&123.3M&60.6 (1.9$\uparrow$) &63.3 (2.4$\uparrow$)\\
    \hline
    UPer-Swin \cite{liu2021swin}&Swin-Base&119.9M&58.4&61.1\\
    UPer-Swin + \textbf{Ours}&Swin-Base&122.5M&\textbf{61.2 (2.8$\uparrow$)}&\textbf{64.2 (3.1$\uparrow$)}\\
    \hline
\end{tabular}
\end{center}
\vspace{-2mm}
\end{table*}
\begin{table*}[tbp]
\setlength{\abovecaptionskip}{0cm}
\setlength{\belowcaptionskip}{-0.2cm}
\caption{Comparison results of day-time and night-time segmentation in terms of mIoU. We evaluated methods on three datasets: the NightCity dataset (denoted as N),  the NightCity-fine dataset (denoted as N-fine), and the cityscapes dataset (denoted as C).  The columns mIoU (C) and mIoU (N) indicate the dataset on which the methods are evaluated. Results marked with * denote metrics reported in the original article. The best score is indicated in \textbf{bold}.}
\label{tab:n_c}
\begin{center}
\renewcommand\arraystretch{1.04}
\begin{tabular}{lcccccc}
\hline
\multirow{2}{*}{Method} & \multirow{2}{*}{Backbone} & \multirow{2}{*}{Parameters} & \multicolumn{2}{c}{trained on C + N} & \multicolumn{2}{c}{trained on C + N-fine} \\
\cline{4-7}
&&&mIoU (N)&mIoU (C)&mIoU (N-fine)&mIoU (C)\\
\hline
NightCity \cite{tan2021night}&Res101&84.6M&\ \ 53.9$^*$&\ \ 76.9$^*$&55.6&76.5\\
NightLab \cite{deng2022nightlab}&Swin-Base&242.4M&60.2&77.1&62.6&77.2\\
\hline
HRNetV2 \cite{HRNetV2}&HRNet-W48&65.8M&58.6&75.0&61.0&76.0\\
HRNetV2 + \textbf{Ours}&HRNet-W48&72.8M&59.6 (1.0$\uparrow$)&75.7&62.9 (1.9$\uparrow$)&77.1\\
\hline
DLV3P \cite{chen2018deeplabv3p}&Res101&60.1M&59.0&73.6&60.9&74.1\\
DLV3P + \textbf{Ours}&Res101&66.3M&59.9 (0.9$\uparrow$)&75.2&62.2 (1.3$\uparrow$)&75.8\\
\hline
UPer-ConvNeXt \cite{liu2022convnet}&ConvNeXt-Base&120.7M&60.1&76.7&61.9&77.9\\
UPer-ConvNeXt + \textbf{Ours}&ConvNeXt-Base&127.7M&61.8 (1.7$\uparrow$)&78.1&64.2 (2.3$\uparrow$)&78.8\\
\hline
UPer-Swin \cite{liu2021swin}&Swin-Base&119.9M&59.7&76.0&62.0&77.9\\
UPer-Swin + \textbf{Ours}&Swin-Base&126.9M&\textbf{63.3 (3.6$\uparrow$)}&\textbf{78.3}&\textbf{64.8 (2.8$\uparrow$)}&\textbf{79.2}\\
\hline
\end{tabular}
\end{center}
\vspace{-5mm}
\end{table*}
\begin{table*}[tbp]
\setlength{\abovecaptionskip}{0cm}
\setlength{\belowcaptionskip}{-0.2cm}
\caption{Comparison results of BDD100K in terms of mIoU. B-N denotes the BDD100K-night dataset, and B-D denotes the BDD100K-day dataset. Results marked with * denote those reported in the original article. The best score is indicated in 
\textbf{bold}.}
\label{tab:bdd100k}
\begin{center}
\begin{tabular}{lcccc}
\hline
\multirow{2}{*}{Method} & \multirow{2}{*}{Backbone} & \multirow{2}{*}{Parameters} & \multicolumn{2}{c}{mIoU (BDD100K-night)}\\
\cline{4-5}
&&&trained on B-N&trained on B-D and B-N\\
\hline
NightCity \cite{tan2021night}&Res101&84.6M&28.4&39.7\\
NightLab \cite{deng2022nightlab}&Swin-Base&242.4M&~~35.4$^*$&~~50.4$^*$\\
\hline
HRNetV2 \cite{HRNetV2}&HRNet-W48&65.8M&29.9&44.3\\
DeeplabV3+ \cite{chen2018deeplabv3p}&Res101&60.1M&30.1&43.4\\
UPer-ConvNeXt \cite{liu2022convnet}&ConvNeXt-Base&120.7M&31.6&47.4\\
\hline
UPer-Swin \cite{liu2021swin} &Swin-Base&119.9M&31.7&48.0\\
UPer-Swin + \textbf{Ours}&Swin-Base&122.5M&\textbf{36.9 (4.2$\uparrow$)}&\textbf{53.1 (5.1$\uparrow$)}\\
\hline
\end{tabular}
\end{center}
\vspace{-4mm}
\end{table*}

To implement disentanglement with entanglement guidance, we recombined the illumination and reflectance components after disentanglement according to the Retinex theory, producing two new images with different illumination. 
New synthetic images then are disentangled again using a weight-sharing disentanglement module, as shown in:
\begin{align}
    \label{eq:3}
    R_j^{s},I_k^{s}=M_{dis} (R_j\odot I_k^\prime ) \notag
    \\
    R_k^{s},I_j^{s}=M_{dis} (R_k\odot I_j^\prime ).
\end{align}
The disentanglement loss is defined as the pixel loss of the reflectance extracted from both original and synthetic images. This loss function constrains the model to learn the consistency of reflectance under different lightness conditions, which can be expressed as:
\begin{equation}
\label{eq:4}
\begin{split}
    \mathcal{L}_{disentangle}={\Vert R_j^{s}-R_j\Vert}_1+{\Vert R_k^{s}-R_k\Vert}_1
    \\
    +{\Vert I_j^{s}-I_j^\prime \Vert}_1+{\Vert I_k^{s}-I_k^\prime \Vert}_1.
\end{split}
\end{equation}

Furthermore, for each pair of disentangled results, we apply the retinex loss to ensure that the organization of reflectance and illumination aligns with the Retinex theory. The retinex loss can be formally described as:
\begin{equation}
    \label{eq:5}
    \mathcal{L}_{retinex}={\Vert I\odot R-X \Vert}_2,
\end{equation}
where X denotes input image. For each generated illumination, a smooth loss is applied to ensure its smoothness, which can be written as:
\begin{equation}
    \label{eq:6}
    \mathcal{L}_{smooth}={\Vert\nabla I\odot \exp{(-\lambda_g\nabla R)}\Vert}_2,
\end{equation}
where $\nabla $ denotes gradient operator. $\lambda_g$ balances the strength of structure-awareness, which is set to 10 as \cite{wei2018deep}. $I$ and $R$ are illumination and reflectance, respectively.

Although the disentanglement module can validly estimate the illumination with the aforementioned losses, it is subject to broad constraints for reflectance, leading to blurring and distortion in the generated reflectance. To  obtain reflectance with clear semantics and boundaries, we introduce the semantic disentanglement loss for generating superior reflectance, is defined as: 
\begin{align}
\label{eq:7}
    &\mathcal{L}_{SeDe}\!=\!\mathcal{L}_{seg} (M_{seg}(R_j\!,\!I_j),Y_j )\!+\!\mathcal{L}_{seg} (M_{seg}(R_k,I_k)\!,\!Y_k) \nonumber
    \\
    &\!+\!\mathcal{L}_{seg} (M_{seg}(R_j^{s},I_j)\!,\!Y_j )\!+\!\mathcal{L}_{seg} (M_{seg}(R_k^{s},I_k),Y_k), 
\end{align}
where $M_{seg}$ refers to segmentation module, $Y_j$ and $Y_k$ are corresponding semantic annotations. Our scheme on how to combine illumination and reflectance for better segmentation will be explained in the next section.

\subsection{Illumination-Aware Parser}
\label{sec:3.3}
The semantic characteristics of certain object categories can be more distinguishable in the illumination features rather than reflectance features. For instance, the semantics of artificial light sources are highly correlated with their illumination. Hence, combining both reflectance and illumination features can lead to more accurate segmentation results compared to using reflectance features alone.

To exploit the semantic features present in illumination, we propose the Illumination-Aware Parser (IAParser), as depicted in Fig.~\ref{fig:overview} (b). For reflectance, the feature extraction module can be an existing semantic segmentation network, (\eg, UPer-Swin \cite{liu2021swin} and DeepLabv3+ \cite{chen2018deeplabv3p}). For illumination, IAParser employs a simple pyramid pooling module \cite{zhao2017pyramid} to extract illumination features. Since not all illumination features are closely associated with semantic information, we use a convolution layer to compute an attention mask that quantitatively evaluates the pixel-wise richness of semantic information contained in the illumination:
\begin{equation}
    \label{eq:8}
    A_{mask}=\sigma(W_a\otimes (W_{ai}\otimes F_{ill}+W_{ar}\otimes F_{ref} )).
\end{equation}
Here, $F_{ill}=M_{ill} (I)$, and $F_{ref}=M_{ref} (R)$. $W$ with different subscripts denotes the learnable weights for the convolution, $\otimes$ denotes the convolution operation, $\sigma$ refers to sigmoid function, and $A_{att}$ is the attention mask that indicates the pixel-wise richness of semantic information in illumination. The final result is obtained by combining $F_{ill}$ and $F_{ref}$ by $M_{att}$:
\begin{equation}
    \label{eq:9}
    \Tilde{Y}=W_{cls}\otimes ((1-A_{mask})\odot F_{ref}+A_{mask}\odot F_{ill}),
\end{equation}
where $W_{cls}$ is a learnable weights for the convlution, which outputs the final prediction $\Tilde{Y}$.

The aforementioned steps implicitly guide the segmentation model to learn the relationship between illumination and semantics. Furthermore, adding explicit constraints as guidance can result in more precise and effective learning of this relationship. To this end, we leverage semantic annotation as a means of adding explicit constraints. Specifically, we use the cross-entropy loss to explicitly guide the generation of clear semantics in $F_{ill}$. The corresponding illumination segmentation loss is defined as follows:
\begin{equation}
    \label{eq:10}
    \mathcal{L}_{segill}=\mathcal{L}_{ce} (W_{cls}\otimes F_{ill},Y).
\end{equation}

In summary, The overall loss function for the IAParser results in the:
\begin{equation}
    \label{eq:12}
    \mathcal{L}_{seg}=\lambda_i \mathcal{L}_{segill}+\mathcal{L}_{ce} (\Tilde{Y},Y),
\end{equation}
where $\lambda_i$ are the hyperparameters for the segmentation loss. This loss function is applied for every result of the $M_{dis}$ output, as described in Eq.~(\ref{eq:7}).
\section{Experiments}
\label{sec:4}
\subsection{Datasets}
\begin{figure}[htbp]
    \centering
    \includegraphics[width=1.0\linewidth]{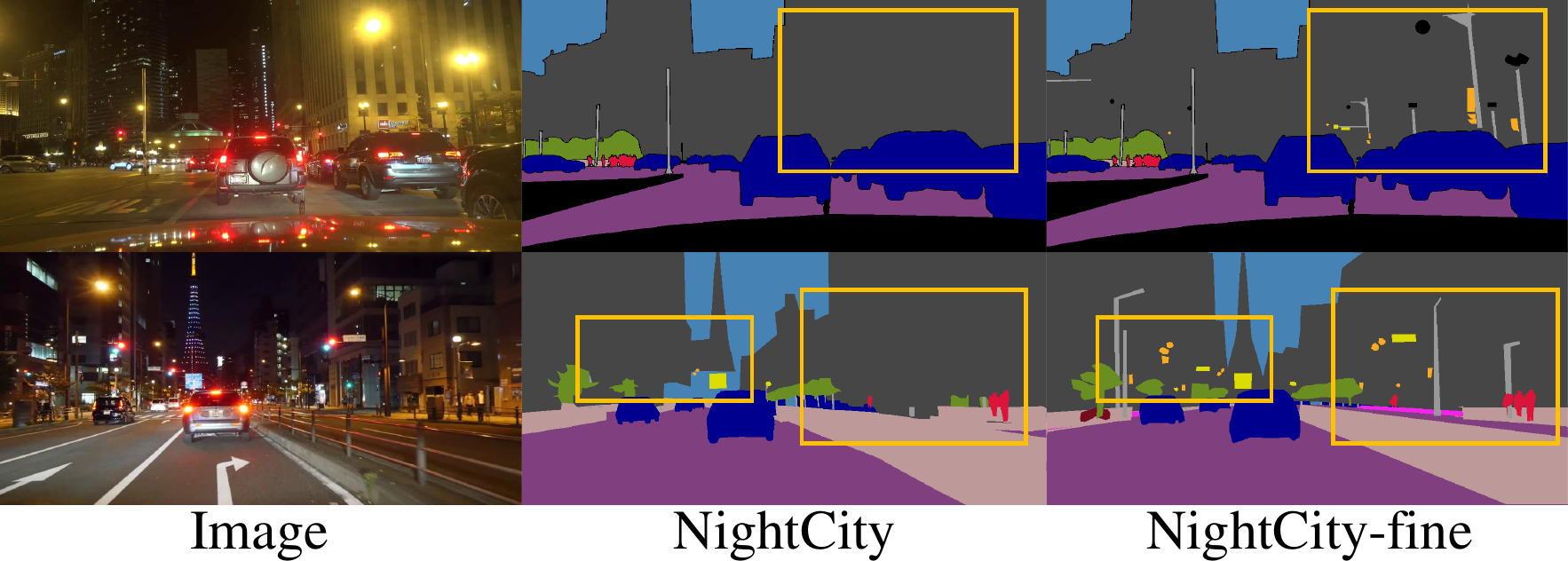}    
    \caption{Samples of modification between NightCity-fine and NightCity datasets.}
    \vspace{-2mm}
    \label{fig:dataset}
\end{figure}
\begin{figure*}[htbp]
    \centering
    \includegraphics[width=1.0\linewidth]{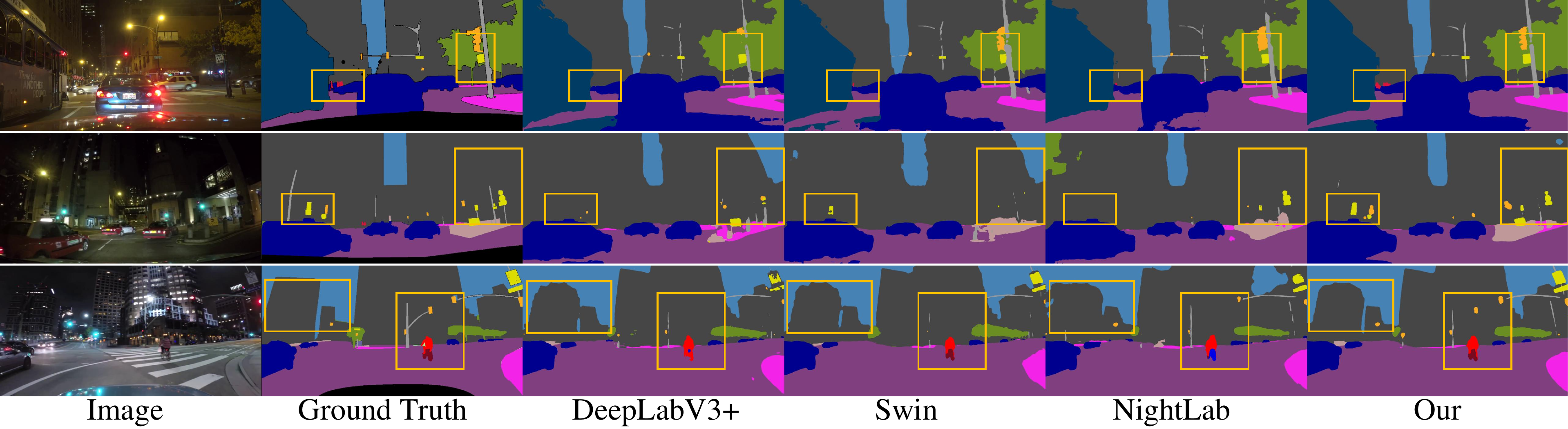}
    \caption{Qualitative comparison of the proposed method with state-of-the-art methods on the NightCity-fine dataset.}
    \label{fig:compare_seg}
\end{figure*}
\begin{figure}[htbp]
    \centering
    \includegraphics[width=1.0\linewidth]{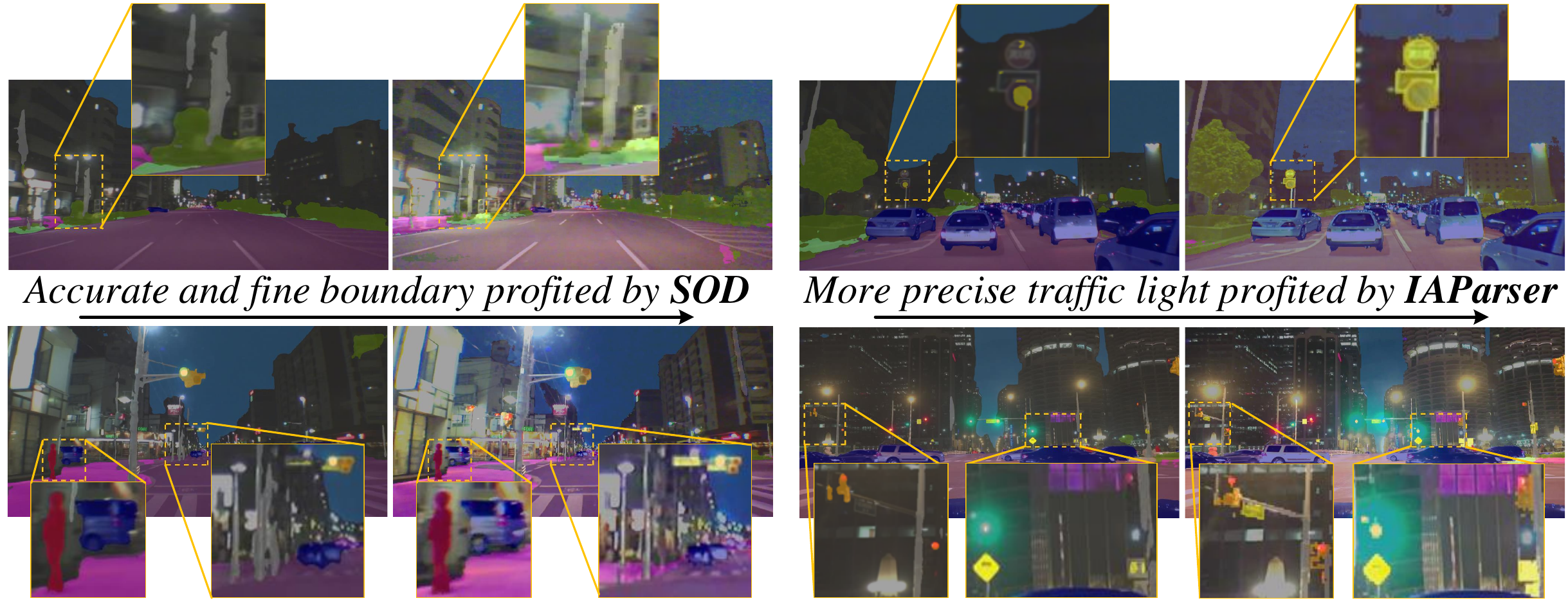}
    \caption{A visual demonstration of performance improvement achieved by SOD and IAParser, compared to the baseline method of Swin. For the same image, the left shows the results of the baseline, while the right shows the results of the proposed method.}
    \vspace{-2mm}
    \label{fig:ablation_visual}
\end{figure}

\noindent\textbf{NightCity-fine}. The original NightCity \cite{tan2021night,deng2022nightlab} is the largest available night-time semantic segmentation dataset, which contains 2998 images with a resolution of 1024x512 for training and 1299 images for validation. All of these samples have pixel-level annotations. However, it is observed that there are some obvious annotation errors that damage the effectiveness of NightCity, as shown in the missing labels and mislabeled regions in Fig.~\ref{fig:dataset}. To address these problems, we propose NightCity-fine, a refined night-time semantic segmentation dataset, in which the unreasonable annotations have been carefully modified in both the training and validation sets, which can be observed in Fig.~\ref{fig:dataset}. More examples of modification can be found in the \textbf{supplementary material}. A total of 2554 label maps have been rectified, and a detailed analysis of the performance that NightCity-fine improves is presented in Tab.~\ref{tab:dataset_ablation}.

\noindent\textbf{Cityscapes} \cite{cordts2016cityscapes}. This is an autonomous driving dataset, which is captured from 50 different cities with 19 semantic classes in the day-time scene. Cityscapes dataset contains 2975 for training and 500 images for validation, both with a resolution of 2048x1024.

\noindent\textbf{BDD100K}. Following the setting in previous work \cite{tan2021night,deng2022nightlab}, we conduct a supplementary experiment using the subset of the BDD100K dataset \cite{yu2020bdd100k}, denoted as BDD100K-night, which includes 314 night-time images for the training set and 31 images for validation. The complementary dataset of BDD100K-night is referred to as BDD100K-day.


\subsection{Implementation details}
\noindent\textbf{Network Architecture}. Our implementation is based on the mmsegmentation framework \cite{mmseg2020}. For the disentanglement model $M_{dis}$, we use a classical encoder-decoder CNN that generates an illumination and reflectance image with the same size as the input. For the illumination segmentation model $M_{ill}$, we use a pyramid pooling module \cite{zhao2017pyramid} for extracting the features. The reflectance segmentation model $M_{ref}$ can be replaced with existing segmentation networks, such as UPer-Swin \cite{liu2021swin} and DeepLabv3+ \cite{chen2018deeplabv3p}.

\noindent\textbf{Training}. During the training process, we adhere to the pipeline recommended by the MMSegmentation \cite{mmseg2020}. This includes mean subtraction, random resizing, and random flipping. Specifically, we resize the images to 512x1024 and then apply random cropping for large images or padding for small images to achieve a fixed size of 256x512, which is used for both Nightcity-fine and Cityscapes datasets. To optimize our network, We employ the AdamW optimizer \cite{adamw} with a base learning rate of 6e-5. The training is carried out for 80000 iterations with a batch size of 8.

\noindent\textbf{Inference}. To handle varying image sizes during inference, we use re-scaled versions of the input image with scaling factors of [0.5, 0.75, 1.0, 1.25, 1.5, 1.75]. We also perform left-right flipping and average the predictions across all augmented versions of the input.

\subsection{Comparison with state-of-the-art Methods}
In this section, we comprehensively evaluate the performance of our method on three datasets: NightCity, NightCity-fine, and BDD100K-night. We compare our method with state-of-the-art methods on the night-time scenes using the NightCity-fine dataset. Moreover, we also evaluate the full-time segmentation performance (both day-time and night-time) by using the NightCity-fine and Cityscapes datasets. For each method, we train them using the hyper-parameters reported in their papers.

\noindent\textbf{NightCity-fine}. Tab.~\ref{tab:nightcity} reports the performance results on the validation set of NightCity and NightCity-fine datasets. Compared to recent methods designed for night-time segmentation, our proposed method achieves a state-of-the-art mIoU score of 61.2\% on NightCity and 64.2\% on NightCity-fine, outperforming existing methods by a significant margin. Additionally, the errors in the original NightCity dataset limit the performance improvement of our proposed method (from 58.4\% to 61.2\%). However, this limit is lifted on the NightCity-fine dataset, resulting in a performance improvement from 61.1\% to 64.2\%. The qualitative results presented in Fig.~\ref{fig:compare_seg} demonstrate that our method can better recognize multiple categories, such as traffic light, traffic sign, and person categories, and generate more precise boundaries.

\noindent\textbf{NightCity-fine and Cityscapes}. We perform experiments under the full-time segmentation task by training networks on both NightCity-fine and Cityscapes jointly. As shown in Tab.~\ref{tab:n_c}, our methods obtain a consistent performance improvement on the full-time scene, outperforming the previous SOTA methods over 2\%, achieving the best performance separately and jointly.

\noindent\textbf{BDD100K}. Tab.~\ref{tab:bdd100k} displays the comparison on BDD100K-day and BDD100K-night. Although BDD100K-night has only 314 images for training, our method still achieves an impressive performance with an mIoU score of 36.9\%, outperforming the SOTA methods by a significant margin.
\begin{table}[tbp]
\setlength{\abovecaptionskip}{0cm}
\setlength{\belowcaptionskip}{-0.3cm}
\caption{Ablation on the key components of our method on NightCity-fine dataset. The best scores are indicated in \textbf{bold}.}
\label{tab:main_ablation}
\vspace{-4mm}
\begin{center}
\scalebox{0.95}{
\renewcommand\arraystretch{1.1}
\setlength\tabcolsep{2.0pt}{
\begin{tabular}{c|cccc|cc}
\hline
 & \multicolumn{4}{c|}{Components} & mIoU & Gain \\ \hline
 & \multicolumn{4}{c|}{Baseline} & 61.1 &  \\ \hline
 & \multicolumn{4}{c|}{HN+grayscale conversion} & 61.6 & +0.5 \\ \hline
 & \multicolumn{4}{c|}{CLAHE+grayscale conversion} & 62.0 & +0.9 \\ \hline
\multirow{6}{*}{SOD} & $\mathcal{L}_{distangle}$ & $\mathcal{L}_{retinex}$ & $\mathcal{L}_{smooth}$ & $\mathcal{L}_{SeDe}$ & mIoU & Gain \\ \cline{2-7} 
& \checkmark &  &  &  & 61.4 & +0.3 \\
& & \checkmark &  & & 61.3 & +0.2 \\
& &  & \checkmark & & 61.3 & +0.2 \\
 & \checkmark &\checkmark & \checkmark &  & 62.5 & +1.4 \\
 & \checkmark & \checkmark & \checkmark & \checkmark & \textbf{63.7} & \textbf{+2.6} \\ \cline{2-7}
 & \multicolumn{4}{c|}{Without disturb operation} & 62.5 & +1.4 \\ \hline
\multirow{3}{*}{IAParser} & \multicolumn{2}{c}{$\mathcal{L}_{seg}$} & \multicolumn{2}{c|}{$\mathcal{L}_{segill}$} & mIoU & Gain \\ \cline{2-7}
 & \multicolumn{2}{c}{\checkmark} & \multicolumn{2}{c|}{} & \multicolumn{1}{c}{63.7} & \multicolumn{1}{c}{+2.6} \\
 & \multicolumn{2}{c}{\checkmark} & \multicolumn{2}{c|}{\checkmark} & \multicolumn{1}{c}{\textbf{64.2}} & \multicolumn{1}{c}{\textbf{+3.1}} \\ \hline
\end{tabular}
}
}
\vspace{-4mm}
\end{center}
\end{table}
\begin{table}[tbp]
\setlength{\abovecaptionskip}{0cm}
\setlength{\belowcaptionskip}{-0.2cm}
\caption{Performance comparison (mIoU) of the disentanglement model with different depths. The best score is indicated in \textbf{bold}.}
\label{tab:sgid_ablation}
\vspace{-4mm}
    \begin{center}
    \scalebox{0.8}
    {
\renewcommand\arraystretch{1.04}
\begin{tabular}{ll|ccccc}
\hline
\multicolumn{2}{l|}{\begin{tabular}[c]{@{}l@{}}Disentangle\\ Model\end{tabular}} & \begin{tabular}[c]{@{}c@{}}None\\ (baseline)\end{tabular} & Small & Base & Large & Huge \\ \hline
\multicolumn{2}{l|}{Parameters} & 121M & +0.2M & +1.5M & +5.9M & +26.8M \\ \hline
\multicolumn{1}{l|}{\multirow{2}{*}{mIoU}} & N-fine & 61.1 & 64.1 &\textbf{64.2} & \textbf{64.2} & \textbf{64.2} \\
\multicolumn{1}{l|}{} & N-fine+ C & 62.0 & 64.2& 64.7 & \textbf{64.8} & \textbf{64.8} \\ \hline
\end{tabular}
    }
    \end{center}    
\vspace{-4mm}
\end{table}

\subsection{Ablation studies}
In ablation studies, UPer-Swin \cite{liu2021swin} is used as $M_{ref}$ due to its strong performance.

\noindent\textbf{Contribution of the losses}. To investigate the influence of different loss functions in our proposed SOD and IAParser, we conduct an ablation study on the critical losses using the NightCity-fine dataset. As presented in Tab.~\ref{tab:main_ablation}, when the model is trained using only one of the three losses, $\mathcal{L}_{dis}$, $\mathcal{L}_{retinex}$, or $\mathcal{L}_{smooth}$, the insufficient constraints hinder the disentanglement process and resulting in only slight improvement. However, when these losses are combined, the illumination and reflectance can be disentangled more effectively, leading to a significant performance boost. Moreover, incorporating $\mathcal{L}_{SeDe}$ into the disentanglement model produces reflectance with clear semantics and boundaries, resulting in even better performance. Finally, based on SOD, adding $\mathcal{L}_{segill}$ further enhances the performance.

\noindent\textbf{Study on SOD}. We explored various configurations that consists of simple pre-processings methods, such as histogram normalization (HN) or CLAHE for generating reflectance, and grayscale conversion for generating illumination. As presented in Tab.~\ref{tab:main_ablation}, these methods are highly limited in obtaining performance gains
. Moreover, removing disturb operation will also cause score decline (64.2\% $\to$ 62.5\%). As shown in Tab.~\ref{tab:sgid_ablation}, we test the impact of disentanglement models at different depths on SOD. The base model trained on NightCity-fine achieves a strong mIoU of 64.2\% with only 1.5 million parameters, and the large model jointly trained on NightCity-fine and Cityscapes datasets achieves the best mIoU of 64.8\% with 5.9 million parameters. Based on this observation that a deeper disentanglement model can handle the larger scale dataset, we select the base model on NightCity-fine and the large model on NightCity-fine and Cityscapes. Additionally, Fig.~\ref{fig:ablation_visual} visually demonstrates that SOD brings accurate and fine boundaries.

\begin{table}[tbp]
\setlength{\abovecaptionskip}{0cm}
\caption{Ablation study about NightCity-fine dataset. N denotes original NightCity dataset, and N-fine denotes refined NightCity-fine dataset. The best score of same method is indicated in \textbf{bold}.}
\vspace{2mm}
\label{tab:dataset_ablation}
\renewcommand\arraystretch{1.04}
\scalebox{0.95}{
\begin{tabular}{l|llcc}
\hline
Method & Training set & Validation set & mIoU& Gain \\ \hline
\multicolumn{1}{l|}{\multirow{4}{*}{Swin}} & N & N & 58.4&\\
\multicolumn{1}{l|}{} & N-fine & N &  60.0 & +1.6\\
\cline{2-5}
\multicolumn{1}{l|}{} & N & N-fine &  58.8 & \\
\multicolumn{1}{l|}{} & N-fine & N-fine &  \textbf{61.1}  & \textbf{+2.3}\\ \hline
\multirow{4}{*}{Swin+Ours} & N & N & 61.2 &\\
& N-fine & N & 62.5 & +1.3 \\
\cline{2-5}
& N & N-fine & 61.5 & \\
& N-fine & N-fine &  \textbf{64.2} & \textbf{+2.7}\\ \hline
\end{tabular}
\vspace{-6mm}
}

\end{table}
\begin{figure}[tbp]
    \centering
    \includegraphics[width=1.0\linewidth]{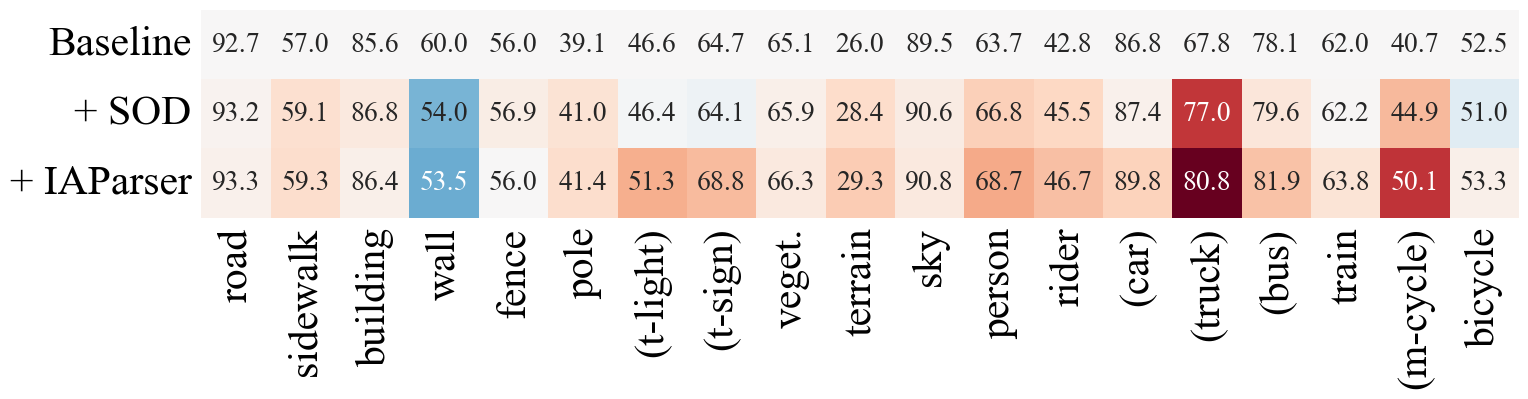}
    \caption{Per-class IoU comparison results on the NightCity-fine dataset. The orange and blue colors denote the IoU gain or degradation to the baseline (Swin). The classes that we observe to relate to light are highlighted in brackets. }
    \label{fig:per_class_iou}
\end{figure}

\noindent\textbf{Study on IAParser}. Fig.~\ref{fig:per_class_iou} reports segmentation results of per-class IoU scores on the NightCity-fine dataset. Thanks to combining the features of reflectance and illumination, our method performs better in categories that are related to lighting or more discriminative in the illumination features, e.g., IoU of 'traffic light' increased from 46.4\% to 51.3\%. Meanwhile, the IoU of categories 'traffic sign', 'car', and 'bus' have improved by at least 2\%. In terms of visual quality, we can observe in Fig.~\ref{fig:ablation_visual} that IAParser makes the more accurate recognition for traffic light and traffic sign.

\noindent\textbf{Study on NightCity-fine}. To verify that our relabeling efforts not only improve the training effect of the training set but also improve the accuracy of the verification of the validation set, we conduct experiments about the performance of both the state-of-the-art methods and our proposed method, as shown in Tab.~\ref{tab:dataset_ablation}.  Compared to the original NightCity dataset, the addition of the relabeled training set results in significant performance gains with the same validation set. Meanwhile, the addition of the relabeled validation set results in more valid performance metrics.

\section{Conclusions}
In this paper, we present a novel night-time semantic segmentation paradigm, \ie disentangle then parse (DTP). Specifically, we come up with a semantic-oriented disentanglement (SOD) framework for night-time scenes, which enables segmentation free from the interference of complicated illumination. Moreover, we propose an illumination-aware parser (IAParser) to leverage the semantic cues embedded in illumination for more precise predictions. DTP can serve as a plug-and-play paradigm to existing methods, which helps these methods achieve superior performance with negligible additional parameters. Furthermore, we refine the largest night-time segmentation dataset NightCity and thus propose the NightCity-fine for more effective training and valid evaluation. 
Extensive experiments across various settings showcase that DTP significantly outperforms state-of-the-art methods, providing a superior night-time segmentation benchmark, together with NightCity-fine.

{
\small
\bibliographystyle{ieee_fullname}
\bibliography{egbib}

\begin{thebibliography}{10}\itemsep=-1pt

\bibitem{night-day-trans-recog}
Asha Anoosheh, Torsten Sattler, Radu Timofte, Marc Pollefeys, and Luc Van~Gool.
\newblock Night-to-day image translation for retrieval-based localization.
\newblock In {\em 2019 International Conference on Robotics and Automation
  (ICRA)}, pages 5958--5964. IEEE, 2019.

\bibitem{Baslamisli_2018_ECCV}
Anil~S. Baslamisli, Thomas~T. Groenestege, Partha Das, Hoang-An Le, Sezer
  Karaoglu, and Theo Gevers.
\newblock Joint learning of intrinsic images and semantic segmentation.
\newblock In {\em Proceedings of the European Conference on Computer Vision
  (ECCV)}, September 2018.

\bibitem{bengio2013representation}
Yoshua Bengio, Aaron Courville, and Pascal Vincent.
\newblock Representation learning: A review and new perspectives.
\newblock {\em IEEE transactions on pattern analysis and machine intelligence},
  35(8):1798--1828, 2013.

\bibitem{bousmalis2017unsupervised}
Konstantinos Bousmalis, Nathan Silberman, David Dohan, Dumitru Erhan, and Dilip
  Krishnan.
\newblock Unsupervised pixel-level domain adaptation with generative
  adversarial networks.
\newblock In {\em Proceedings of the IEEE conference on computer vision and
  pattern recognition}, pages 3722--3731, 2017.

\bibitem{bousmalis2016domain}
Konstantinos Bousmalis, George Trigeorgis, Nathan Silberman, Dilip Krishnan,
  and Dumitru Erhan.
\newblock Domain separation networks.
\newblock {\em Advances in neural information processing systems}, 29, 2016.

\bibitem{cai2018learning-deep-single}
Jianrui Cai, Shuhang Gu, and Lei Zhang.
\newblock Learning a deep single image contrast enhancer from multi-exposure
  images.
\newblock {\em IEEE Transactions on Image Processing}, 27(4):2049--2062, 2018.

\bibitem{SSSIPC}
Huaian Chen, Yi Jin, Guoqiang Jin, Changan Zhu, and Enhong Chen.
\newblock Semisupervised semantic segmentation by improving prediction
  confidence.
\newblock {\em IEEE Transactions on Neural Networks and Learning Systems},
  33(9):4991--5003, 2022.

\bibitem{DDB}
Lin Chen, Zhixiang Wei, Xin Jin, Huaian Chen, Miao Zheng, Kai Chen, and Yi Jin.
\newblock Deliberated domain bridging for domain adaptive semantic
  segmentation.
\newblock In S. Koyejo, S. Mohamed, A. Agarwal, D. Belgrave, K. Cho, and A. Oh,
  editors, {\em Advances in Neural Information Processing Systems}, volume~35,
  pages 15105--15118. Curran Associates, Inc., 2022.

\bibitem{chen2014deeplab}
Liang-Chieh Chen, George Papandreou, Iasonas Kokkinos, Kevin Murphy, and Alan~L
  Yuille.
\newblock Semantic image segmentation with deep convolutional nets and fully
  connected crfs.
\newblock {\em arXiv preprint arXiv:1412.7062}, 2014.

\bibitem{chen2017deeplabv2}
Liang-Chieh Chen, George Papandreou, Iasonas Kokkinos, Kevin Murphy, and Alan~L
  Yuille.
\newblock Deeplab: Semantic image segmentation with deep convolutional nets,
  atrous convolution, and fully connected crfs.
\newblock {\em IEEE transactions on pattern analysis and machine intelligence},
  40(4):834--848, 2017.

\bibitem{chen2017deeplabv3}
Liang-Chieh Chen, George Papandreou, Florian Schroff, and Hartwig Adam.
\newblock Rethinking atrous convolution for semantic image segmentation.
\newblock {\em arXiv preprint arXiv:1706.05587}, 2017.

\bibitem{chen2018deeplabv3p}
Liang-Chieh Chen, Yukun Zhu, George Papandreou, Florian Schroff, and Hartwig
  Adam.
\newblock Encoder-decoder with atrous separable convolution for semantic image
  segmentation.
\newblock In {\em Proceedings of the European conference on computer vision
  (ECCV)}, pages 801--818, 2018.

\bibitem{HANet}
Sungha Choi, Joanne~T Kim, and Jaegul Choo.
\newblock Cars can't fly up in the sky: Improving urban-scene segmentation via
  height-driven attention networks.
\newblock In {\em Proceedings of the IEEE/CVF conference on computer vision and
  pattern recognition}, pages 9373--9383, 2020.

\bibitem{mmseg2020}
MMSegmentation Contributors.
\newblock {MMSegmentation}: Openmmlab semantic segmentation toolbox and
  benchmark.
\newblock \url{https://github.com/open-mmlab/mmsegmentation}, 2020.

\bibitem{cordts2016cityscapes}
Marius Cordts, Mohamed Omran, Sebastian Ramos, Timo Rehfeld, Markus Enzweiler,
  Rodrigo Benenson, Uwe Franke, Stefan Roth, and Bernt Schiele.
\newblock The cityscapes dataset for semantic urban scene understanding.
\newblock In {\em Proceedings of the IEEE conference on computer vision and
  pattern recognition}, pages 3213--3223, 2016.

\bibitem{dark-adaptation}
Dengxin Dai and Luc Van~Gool.
\newblock Dark model adaptation: Semantic image segmentation from daytime to
  nighttime.
\newblock In {\em 2018 21st International Conference on Intelligent
  Transportation Systems (ITSC)}, pages 3819--3824. IEEE, 2018.

\bibitem{deng2022nightlab}
Xueqing Deng, Peng Wang, Xiaochen Lian, and Shawn Newsam.
\newblock Nightlab: A dual-level architecture with hardness detection for
  segmentation at night.
\newblock In {\em Proceedings of the IEEE/CVF Conference on Computer Vision and
  Pattern Recognition}, pages 16938--16948, 2022.

\bibitem{dosovitskiy2020image}
Alexey Dosovitskiy, Lucas Beyer, Alexander Kolesnikov, Dirk Weissenborn,
  Xiaohua Zhai, Thomas Unterthiner, Mostafa Dehghani, Matthias Minderer, Georg
  Heigold, Sylvain Gelly, et~al.
\newblock An image is worth 16x16 words: Transformers for image recognition at
  scale.
\newblock {\em arXiv preprint arXiv:2010.11929}, 2020.

\bibitem{DANet}
Jun Fu, Jing Liu, Haijie Tian, Yong Li, Yongjun Bao, Zhiwei Fang, and Hanqing
  Lu.
\newblock Dual attention network for scene segmentation.
\newblock In {\em Proceedings of the IEEE/CVF conference on computer vision and
  pattern recognition}, pages 3146--3154, 2019.

\bibitem{ganin2015unsupervised}
Yaroslav Ganin and Victor Lempitsky.
\newblock Unsupervised domain adaptation by backpropagation.
\newblock In {\em International conference on machine learning}, pages
  1180--1189. PMLR, 2015.

\bibitem{goodfellow2020generative}
Ian Goodfellow, Jean Pouget-Abadie, Mehdi Mirza, Bing Xu, David Warde-Farley,
  Sherjil Ozair, Aaron Courville, and Yoshua Bengio.
\newblock Generative adversarial networks.
\newblock {\em Communications of the ACM}, 63(11):139--144, 2020.

\bibitem{jiang2021switched}
Zhuqing Jiang, Haotian Li, Liangjie Liu, Aidong Men, and Haiying Wang.
\newblock A switched view of retinex: Deep self-regularized low-light image
  enhancement.
\newblock {\em Neurocomputing}, 454:361--372, 2021.

\bibitem{land1977retinex}
Edwin~H Land.
\newblock The retinex theory of color vision.
\newblock {\em Scientific american}, 237(6):108--129, 1977.

\bibitem{GPS-GLASS}
Hongjae Lee, Changwoo Han, and Seung-Won Jung.
\newblock Gps-glass: Learning nighttime semantic segmentation using daytime
  video and gps data.
\newblock {\em arXiv preprint arXiv:2207.13297}, 2022.

\bibitem{lee2020bright-channel-prior}
Hunsang Lee, Kwanghoon Sohn, and Dongbo Min.
\newblock Unsupervised low-light image enhancement using bright channel prior.
\newblock {\em IEEE Signal Processing Letters}, 27:251--255, 2020.

\bibitem{lee2020drit++}
Hsin-Ying Lee, Hung-Yu Tseng, Qi Mao, Jia-Bin Huang, Yu-Ding Lu, Maneesh Singh,
  and Ming-Hsuan Yang.
\newblock Drit++: Diverse image-to-image translation via disentangled
  representations.
\newblock {\em International Journal of Computer Vision}, 128:2402--2417, 2020.

\bibitem{lin2017refinenet}
Guosheng Lin, Anton Milan, Chunhua Shen, and Ian Reid.
\newblock Refinenet: Multi-path refinement networks for high-resolution
  semantic segmentation.
\newblock In {\em Proceedings of the IEEE conference on computer vision and
  pattern recognition}, pages 1925--1934, 2017.

\bibitem{liu2018detach}
Yen-Cheng Liu, Yu-Ying Yeh, Tzu-Chien Fu, Sheng-De Wang, Wei-Chen Chiu, and
  Yu-Chiang~Frank Wang.
\newblock Detach and adapt: Learning cross-domain disentangled deep
  representation.
\newblock In {\em Proceedings of the IEEE Conference on Computer Vision and
  Pattern Recognition}, pages 8867--8876, 2018.

\bibitem{liu2021swin}
Ze Liu, Yutong Lin, Yue Cao, Han Hu, Yixuan Wei, Zheng Zhang, Stephen Lin, and
  Baining Guo.
\newblock Swin transformer: Hierarchical vision transformer using shifted
  windows.
\newblock In {\em Proceedings of the IEEE/CVF international conference on
  computer vision}, pages 10012--10022, 2021.

\bibitem{liu2022convnet}
Zhuang Liu, Hanzi Mao, Chao-Yuan Wu, Christoph Feichtenhofer, Trevor Darrell,
  and Saining Xie.
\newblock A convnet for the 2020s.
\newblock In {\em Proceedings of the IEEE/CVF Conference on Computer Vision and
  Pattern Recognition}, pages 11976--11986, 2022.

\bibitem{long2015fully}
Jonathan Long, Evan Shelhamer, and Trevor Darrell.
\newblock Fully convolutional networks for semantic segmentation.
\newblock In {\em Proceedings of the IEEE conference on computer vision and
  pattern recognition}, pages 3431--3440, 2015.

\bibitem{adamw}
Ilya Loshchilov and Frank Hutter.
\newblock Decoupled weight decay regularization.
\newblock {\em arXiv preprint arXiv:1711.05101}, 2017.

\bibitem{ma2021retinexgan}
Tian Ma, Ming Guo, Zhenhua Yu, Yanping Chen, Xincheng Ren, Runtao Xi, Yuancheng
  Li, and Xinlei Zhou.
\newblock Retinexgan: Unsupervised low-light enhancement with two-layer
  convolutional decomposition networks.
\newblock {\em IEEE Access}, 9:56539--56550, 2021.

\bibitem{mathieu2016disentangling}
Michael~F Mathieu, Junbo~Jake Zhao, Junbo Zhao, Aditya Ramesh, Pablo
  Sprechmann, and Yann LeCun.
\newblock Disentangling factors of variation in deep representation using
  adversarial training.
\newblock {\em Advances in neural information processing systems}, 29, 2016.

\bibitem{nayar1996reflectance}
Shree~K Nayar and Ruud~M Bolle.
\newblock Reflectance based object recognition.
\newblock {\em International journal of computer vision}, 17:219--240, 1996.

\bibitem{bridging-day-night}
Eduardo Romera, Luis~M Bergasa, Kailun Yang, Jose~M Alvarez, and Rafael Barea.
\newblock Bridging the day and night domain gap for semantic segmentation.
\newblock In {\em 2019 IEEE Intelligent Vehicles Symposium (IV)}, pages
  1312--1318. IEEE, 2019.

\bibitem{unet}
Olaf Ronneberger, Philipp Fischer, and Thomas Brox.
\newblock U-net: Convolutional networks for biomedical image segmentation.
\newblock In {\em Medical Image Computing and Computer-Assisted
  Intervention--MICCAI 2015: 18th International Conference, Munich, Germany,
  October 5-9, 2015, Proceedings, Part III 18}, pages 234--241. Springer, 2015.

\bibitem{sakaridis2019guided}
Christos Sakaridis, Dengxin Dai, and Luc~Van Gool.
\newblock Guided curriculum model adaptation and uncertainty-aware evaluation
  for semantic nighttime image segmentation.
\newblock In {\em Proceedings of the IEEE/CVF International Conference on
  Computer Vision}, pages 7374--7383, 2019.

\bibitem{night-day-trans-dec}
Mark Schutera, Mostafa Hussein, Jochen Abhau, Ralf Mikut, and Markus Reischl.
\newblock Night-to-day: Online image-to-image translation for object detection
  within autonomous driving by night.
\newblock {\em IEEE Transactions on Intelligent Vehicles}, 6(3):480--489, 2020.

\bibitem{strudel2021segmenter}
Robin Strudel, Ricardo Garcia, Ivan Laptev, and Cordelia Schmid.
\newblock Segmenter: Transformer for semantic segmentation.
\newblock In {\em Proceedings of the IEEE/CVF international conference on
  computer vision}, pages 7262--7272, 2021.

\bibitem{HRNetV2}
Ke Sun, Y Zhao, B Jiang, T Cheng, B Xiao, D Liu, Y Mu, X Wang, W Liu, and J
  Wang.
\newblock High-resolution representations for labeling pixels and regions.
  arxiv 2019.
\newblock {\em arXiv preprint arXiv:1904.04514}, 2019.

\bibitem{sun2019see}
Lei Sun, Kaiwei Wang, Kailun Yang, and Kaite Xiang.
\newblock See clearer at night: towards robust nighttime semantic segmentation
  through day-night image conversion.
\newblock In {\em Artificial Intelligence and Machine Learning in Defense
  Applications}, volume 11169, pages 77--89. SPIE, 2019.

\bibitem{tan2021night}
Xin Tan, Ke Xu, Ying Cao, Yiheng Zhang, Lizhuang Ma, and Rynson~WH Lau.
\newblock Night-time scene parsing with a large real dataset.
\newblock {\em IEEE Transactions on Image Processing}, 30:9085--9098, 2021.

\bibitem{van2008visualizing}
Laurens Van~der Maaten and Geoffrey Hinton.
\newblock Visualizing data using t-sne.
\newblock {\em Journal of machine learning research}, 9(11), 2008.

\bibitem{Non-Local}
Xiaolong Wang, Ross Girshick, Abhinav Gupta, and Kaiming He.
\newblock Non-local neural networks.
\newblock In {\em Proceedings of the IEEE conference on computer vision and
  pattern recognition}, pages 7794--7803, 2018.

\bibitem{wei2018deep}
Chen Wei, Wenjing Wang, Wenhan Yang, and Jiaying Liu.
\newblock Deep retinex decomposition for low-light enhancement.
\newblock {\em arXiv preprint arXiv:1808.04560}, 2018.

\bibitem{wu2021dannet}
Xinyi Wu, Zhenyao Wu, Hao Guo, Lili Ju, and Song Wang.
\newblock Dannet: A one-stage domain adaptation network for unsupervised
  nighttime semantic segmentation.
\newblock In {\em Proceedings of the IEEE/CVF Conference on Computer Vision and
  Pattern Recognition}, pages 15769--15778, 2021.

\bibitem{FDLNet}
Zhifeng Xie, Sen Wang, Ke Xu, Zhizhong Zhang, Xin Tan, Yuan Xie, and Lizhuang
  Ma.
\newblock Boosting night-time scene parsing with learnable frequency.
\newblock {\em IEEE Transactions on Image Processing}, 32:2386--2398, 2023.

\bibitem{Lawin-Transformer}
Haotian Yan, Chuang Zhang, and Ming Wu.
\newblock Lawin transformer: Improving semantic segmentation transformer with
  multi-scale representations via large window attention.
\newblock {\em arXiv preprint arXiv:2201.01615}, 2022.

\bibitem{yu2020bdd100k}
Fisher Yu, Haofeng Chen, Xin Wang, Wenqi Xian, Yingying Chen, Fangchen Liu,
  Vashisht Madhavan, and Trevor Darrell.
\newblock Bdd100k: A diverse driving dataset for heterogeneous multitask
  learning.
\newblock In {\em Proceedings of the IEEE/CVF conference on computer vision and
  pattern recognition}, pages 2636--2645, 2020.

\bibitem{yu2015multi}
Fisher Yu and Vladlen Koltun.
\newblock Multi-scale context aggregation by dilated convolutions.
\newblock {\em arXiv preprint arXiv:1511.07122}, 2015.

\bibitem{OCRNet}
Yuhui Yuan, Xilin Chen, and Jingdong Wang.
\newblock Object-contextual representations for semantic segmentation.
\newblock In {\em Computer Vision--ECCV 2020: 16th European Conference,
  Glasgow, UK, August 23--28, 2020, Proceedings, Part VI 16}, pages 173--190.
  Springer, 2020.

\bibitem{OCNet}
Yuhui Yuan, Lang Huang, Jianyuan Guo, Chao Zhang, Xilin Chen, and Jingdong
  Wang.
\newblock Ocnet: Object context network for scene parsing.
\newblock {\em arXiv preprint arXiv:1809.00916}, 2018.

\bibitem{zhao2017pyramid}
Hengshuang Zhao, Jianping Shi, Xiaojuan Qi, Xiaogang Wang, and Jiaya Jia.
\newblock Pyramid scene parsing network.
\newblock In {\em Proceedings of the IEEE conference on computer vision and
  pattern recognition}, pages 2881--2890, 2017.

\bibitem{PSANet}
Hengshuang Zhao, Yi Zhang, Shu Liu, Jianping Shi, Chen~Change Loy, Dahua Lin,
  and Jiaya Jia.
\newblock Psanet: Point-wise spatial attention network for scene parsing.
\newblock In {\em Proceedings of the European conference on computer vision
  (ECCV)}, pages 267--283, 2018.

\bibitem{zhao2021retinexdip}
Zunjin Zhao, Bangshu Xiong, Lei Wang, Qiaofeng Ou, Lei Yu, and Fa Kuang.
\newblock Retinexdip: A unified deep framework for low-light image enhancement.
\newblock {\em IEEE Transactions on Circuits and Systems for Video Technology},
  32(3):1076--1088, 2021.

\end{thebibliography}
}
\end{document}